\documentclass[]{openmoss}

\usepackage{helvet}

\usepackage{amsmath} 
\usepackage{natbib}
\usepackage{graphicx}
\usepackage{subcaption} 

\usepackage[toc,page,header]{appendix}
\usepackage[utf8]{inputenc} 
\usepackage[T1]{fontenc}    
\usepackage{hyperref}       
\usepackage{url}            
\usepackage{booktabs}       
\usepackage{lmodern}        
\usepackage{amsfonts}       
\usepackage{nicefrac}       
\usepackage{microtype}      
\usepackage{wrapfig}

\usepackage{amssymb}        
\usepackage{titletoc}
\usepackage{minitoc}

\usepackage{array}
\usepackage{etoolbox}

\definecolor{lightblue}{RGB}{200, 230, 255}  
\definecolor{headerblue}{RGB}{150, 200, 255} 

\usepackage{pgfplots}
\usepackage{xcolor}
\usepackage{float}
\usepackage{comment}
\usepackage{multirow}
\usepackage{makecell}
\usepackage{siunitx}
\usepackage{tikz}
\usepackage{pgf-pie}
\usepackage[export]{adjustbox}

\usepackage{ragged2e}
\usepackage{tabularx}
\usepackage{caption}
\usepackage{enumitem}
\usepackage{pifont}
\usepackage[hang,flushmargin]{footmisc}

\usepackage{tcolorbox}
\tcbuselibrary{breakable}
\tcbuselibrary{skins}

\usepackage{tabularx}
\usepackage{listings}

\usepackage{threeparttable}
\usepackage{setspace}
\usepackage[scheme=plain]{ctex} 

\definecolor{MossCyan}{HTML}{82D9FF} 
\definecolor{MossBlue}{HTML}{82B1FF}

\definecolor{ForestGreen}{RGB}{34, 139, 34}
\definecolor{Red}{RGB}{255, 0, 0}

\definecolor{tickG}{rgb}{0.1, 0.588, 0.1}
\definecolor{crossR}{rgb}{0.588, 0.1, 0.1}

\definecolor{frenchblue}{rgb}{0.0, 0.45, 0.73}
\definecolor{babyblue}{rgb}{0.54, 0.81, 0.94}
\definecolor{classicrose}{rgb}{0.98, 0.8, 0.91}
\definecolor{beige}{rgb}{0.96, 0.96, 0.86}
\definecolor{forestgreen}{HTML}{2e7d43}

\definecolor{blue1}{HTML}{91BBE6}
\definecolor{blue2}{HTML}{3F90E0}
\definecolor{blue3}{HTML}{316FAD}

\definecolor{color1}{HTML}{FF9999}
\definecolor{color2}{HTML}{FF6666}
\definecolor{color3}{HTML}{FF3333}
\definecolor{color4}{HTML}{E60000}
\definecolor{color5}{HTML}{B30000}
\definecolor{color6}{HTML}{8CD98C}
\definecolor{color7}{HTML}{53c653}
\definecolor{color8}{HTML}{00B050}
\definecolor{color9}{HTML}{2d862d}
\definecolor{color10}{HTML}{206020}
\definecolor{color11}{HTML}{cca300}

\definecolor{eggshell}{HTML}{EFEBE4}
\definecolor{deltaBg}{RGB}{220,230,255} 

\newcommand{\rowhighlight}{\rowcolor{deltaBg}}

\addto\extrasenglish{

}

\newtcolorbox{promptbox}[2][]{
    colback=white,
    coltext=black,
    arc=3mm,
    boxrule=0.5pt,
    colframe=black!60!white,
    title={#2},
    colbacktitle=black,
    coltitle=white,
    fonttitle=\bfseries,
    top=8pt,
    bottom=8pt,
    left=10pt,
    right=10pt,
    breakable,
    before upper={%
        \linespread{1}\selectfont
        \setlength{\parskip}{1ex plus 0.2ex minus 0.2ex}%
        \setlength{\parindent}{0pt}%
    },
    #1
}

\title{In-Context World Modeling for Robotic Control}

\author{
Siyin Wang$^{1,2}$ \hspace{.3em}
Junhao Shi$^{1,2}$\hspace{.3em}
Senyu Fei$^{2,3}$ \hspace{.1em}
Zhaoyang Fu$^{1}$
\\
\textbf{
Li Ji$^{1,2}$ \hspace{.1em}
Jingjing Gong$^{2,\dagger}$ \hspace{.1em}
Xipeng Qiu$^{1,2,\dagger}$ \hspace{.1em}
}
\\
[1ex]
\normalfont
$^{1}$Fudan University   
$^{2}$Shanghai Innovation Institute  
$^{3}$Tongji University  
\\
\texttt{siyinwang20@fudan.edu.cn} \\
}

\abstract{
Modern Vision-Language-Action (VLA) models often fail to generalize to novel setups, such as altered camera viewpoints or robot morphologies, because they are typically conditioned only on current observations and language instructions. By ignoring the underlying system configuration as a variable, these models implicitly assume a fixed execution context encountered during training, necessitating data-intensive fine-tuning for any new environment. In this work, we introduce In-Context World Modeling (ICWM), a framework that treats system identification as an in-context adaptation problem. ICWM enables robot policies to autonomously infer essential system variables from a short history of self-generated, task-agnostic interactions. Unlike traditional In-Context Learning that uses demonstrations to specify what task to perform, ICWM leverages the context window to understand how the system operates. 
By processing these interactions before task execution, the model implicitly captures the world dynamics of the current system, enabling adaptation to novel configurations without parameter updates.
Extensive experiments in simulation and on real-world robot platforms demonstrate that ICWM significantly outperforms standard VLA baselines on novel camera viewpoints.
}

\begin{document}
\maketitle
\begingroup
\renewcommand{\thefootnote}{\fnsymbol{footnote}}
\footnotetext{$^{\dagger}$Corresponding Authors}
\endgroup



\section{Introduction}

Consider a human operator handed a joystick to control a robot, but with no prior knowledge of the control mapping. Does pushing forward move the robot left, right, or forward? The operator's first instinct is not to attempt the task directly, but to explore: move the joystick randomly, observe the robot's response, and infer the input-output relationship. Within seconds, this brief calibration phase yields an internal model of the system dynamics, enabling the transition from stochastic exploration to goal-directed control.
Crucially, this calibration is entirely self-generated—the operator needs no prior knowledge of the task, no external guidance, and no task-specific experience under the new control mapping.

\begin{figure}[t]
    \centering
    \includegraphics[width=\linewidth]{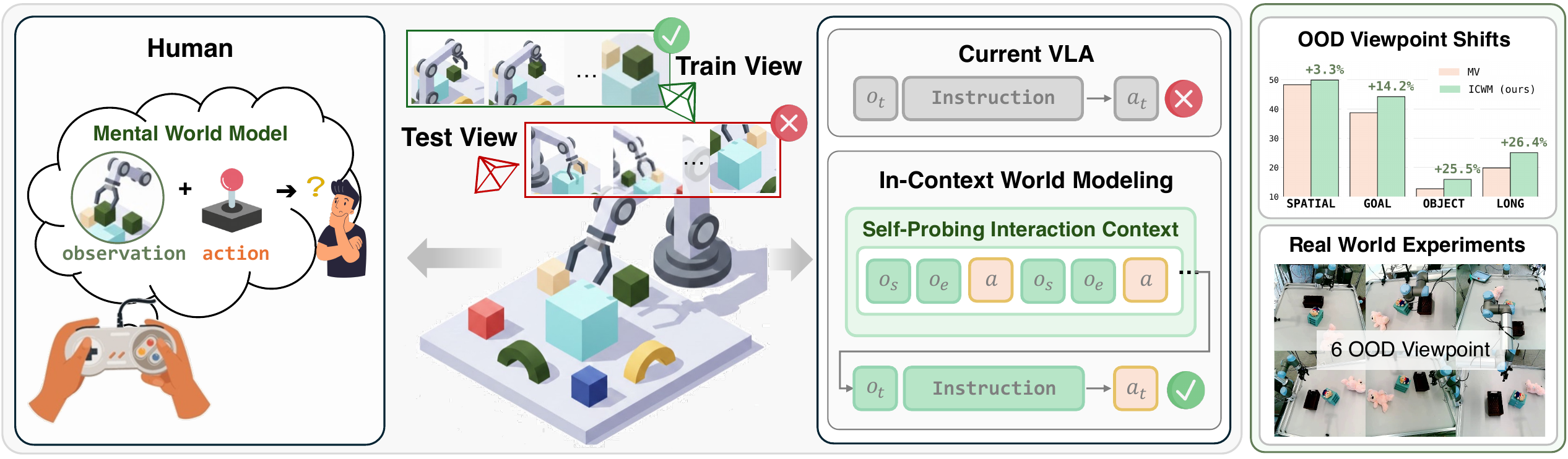}
    \caption{In-Context World Modeling (ICWM). Standard VLA models often fail in novel system configurations due to fixed observation-action assumptions. Similar to how humans explore unfamiliar controls to build a mental world model, ICWM enables robots to autonomously infer system dynamics from self-probing interaction context.}
    \label{fig:intro}
\end{figure}

Modern Vision-Language-Action (VLA) models \cite{rt2,openvla,pi0}, however, lack this calibration ability entirely. The standard formulation $\pi_\theta\left(a_t \mid o_t, l\right)$ conditions only on the current observation and language instruction, treating system configuration $\psi$ (like camera viewpoints, morphology) as a fixed constant absorbed into model parameters during training. 
When deployment conditions deviate from training, the model has no mechanism to recover the correct action-observation correspondence, and performance degrades \cite{Fei2025LIBEROPlusIR}.
Scene-specific fine-tuning remains the prevailing remedy, yet it requires human intervention for every new setup—contradicting the goal of generalist deployment.

We argue this failure is a \textit{system identification} problem: the policy 
lacks knowledge of $\psi$ at test time. This naturally motivates repurposing 
the context window to recover $\psi$ through targeted interaction. However, 
existing In-Context Learning (ICL) methods \cite{Fu2024ICRTII, 
Shah2025MimicDroidIL} treat context as \textit{behavior specification} (what to 
do) rather than \textit{system identification} (how the system operates), and 
still require human-provided demonstrations at test time.

In this work, we propose In-Context World Modeling (ICWM) (\autoref{fig:intro}), which repurposes the context window for system 
identification rather than behavior specification. Before task execution, the robot performs a short sequence of random exploratory movements and records the resulting visual transitions; these self-probed clips are prepended as context, from which the model implicitly recovers the current system configuration and adapts its actions accordingly. 
Surprisingly, the task-agnostic random movements alone provide sufficient context for implicit system identification without any task-specific demonstrations and we show this simple strategy yields consistent improvements over standard VLA training on novel viewpoints across both simulation and real-robot platforms.

In summary, this work makes the following contributions:
\begin{itemize}
    \item We reframe VLA generalization as a test-time system identification problem, identifying the absence of explicit $\psi$ conditioning as an underexplored failure mode that persists across standard training strategies.
    \item We propose In-Context World Modeling (ICWM), which implicitly identifies the system configuration from self-generated, task-agnostic exploratory transitions prepended as context, achieving test-time adaptation without parameter updates or task-specific demonstrations.
    \item We validate ICWM on both simulation benchmarks and real-robot experiments, demonstrating significant improvements over standard VLA training on novel camera viewpoints, with further generalization to semantic scene variations and robot morphological changes.
\end{itemize}

\section{Related Work}

\subsection{In-Context Adaptation for Robotics}
The emergence of large-scale pre-trained models~\cite{Radford2019LanguageMA,Touvron2023LLaMAOA,OpenAI2022ChatGPT} has established In-Context Learning (ICL) as an emergent capability for on-the-fly skill acquisition~\cite{Brown2020LanguageMA,Dong2022ASO}. In robotics, ICL has been explored via in-context imitation: models are prompted with expert trajectories, provided by teleoperation or retrieved from offline buffers, to perform next-token prediction over observation-action sequences~\cite{Fu2024ICRTII,An2025ActionTM,Sridhar2025RICLAI}; complementary approaches leverage human play videos for cross-embodiment transfer~\cite{Shah2025MimicDroidIL,Jain2024Vid2RobotEV}. Despite differences in data modality, all such methods treat context as behavior specification (``what to do''), and critically require human-provided demonstrations at test time. 
A related line of work pursues adaptation through meta-learning. Gradient-based methods~\cite{MAML2017,Liu2019MetaLearningWI} meta-optimize an initialization for rapid test-time fine-tuning, while recurrent meta-RL~\cite{Duan2016RL2FR, Zintgraf2020VariBADAV} encodes interaction history 
to infer \textit{which task} to perform—relying on reward signals unavailable prior to task execution. 
In contrast, ICWM uses task-agnostic self-generated interactions to understand \textit{how the system operates}, requiring no demonstrations, no reward signal, and no parameter updates at test time--a harder and more practical deployment setting.

\subsection{World Modeling for Robotic Control}
World modeling captures environment dynamics to ground an agent's actions in causal state transitions~\cite{Ha2018RecurrentWM,LeCun2022APT, Ding2024UnderstandingWO,wang2025world,wang2026world}. Existing approaches fall into two paradigms: forward models predict future observations either in pixel space~\cite{GR-1, Li2024GRMGLP,Zhao2025CoTVLAVC,Cen2025WorldVLATA} or latent space~\cite{Hafner2019DreamTC,Hafner2020MasteringAW,Wu2022DayDreamerWM, Hafner2023MasteringDD,Zheng2025FLARERL}, while inverse dynamics models abduce actions from observed visual changes~\cite{Du2023LearningUP,Jang2025DreamGenUG,Tian2024PredictiveID,Baker2022VideoP}; some works unify both objectives~\cite{Zhu2025UnifiedWM,Li2025UnifiedVA}. All of these introduce dedicated parameters and training objectives for world modeling. ICWM instead realizes world modeling implicitly: 
the model leverages standard sequence modeling to extract the time-invariant causal structure (e.g., control mappings and camera viewpoints) directly from task-agnostic interaction histories. 
This implicit paradigm introduces zero additional parameters, treating world modeling as an emergent inference capability at test time.


\section{Preliminary and Motivation}
\label{sec:analysis}

\subsection{Standard VLA Formulation and Its Limitation}

We consider a robot manipulation task where a policy $\pi$ maps multimodal observations and language instructions to actions. Let $\mathcal{O}$ be the observation space and $\mathcal{I}$ be the space of natural language instructions. A standard VLA policy $\pi_\theta\left(a_t \mid o_t, l\right)$ processes the current observation $o_t \in \mathcal{O}$ and instruction $l \in \mathcal{I}$ to predict an action $a_t \in \mathcal{A}$. The control task is thus formulated as:
\begin{equation}
\pi_\theta\left(a_t \mid o_t, l\right).
\end{equation}
where the model parameters $\theta$ are optimized on a large-scale dataset $\mathcal{D}$ collected under a set of specific system setups. In practice, $a_t$ is often structured as an action chunk to ensure temporal smoothness and control stability during execution.

This formulation embeds an implicit assumption: the system configuration 
$\psi$—encompassing camera viewpoints, mounting offsets, and the robot's 
morphological properties—is fixed and known. An ideal policy should instead 
condition on $\psi$ explicitly:
\begin{equation}
\pi_\theta^*\left(a_t \mid o_t, l, \psi\right).
\end{equation}
Without $\psi$, training forces the model to marginalize over all 
configurations in the dataset:
\begin{equation}
\pi_\theta\left(a_t \mid o_t, l\right) \approx \int \pi_\theta^*\left(a_t \mid o_t, l, \psi\right) 
p\left(\psi\right) d\psi.
\end{equation}
At deployment, when a specific $\psi'$ is realized, this averaged policy 
lacks the context to correctly interpret the observation-action correspondence, 
leading to degraded performance on novel configurations. This motivates 
explicit recovery of $\psi$ at test time.

\subsection{Interaction Context Enriches Configuration Information}
\label{subsec:theory}

We model robot-environment interaction as a POMDP where the latent state 
decomposes as $s_k = (\psi, \xi_k)$, with $\psi$ being the time-invariant 
system configuration and $\xi_k$ the time-varying scene state. The system 
evolves as:
\begin{equation}
s_0 \xrightarrow{a_1} s_1 \xrightarrow{a_2} \cdots \xrightarrow{a_t} s_t, 
\qquad o_k \sim p\left(o \mid s_k\right).
\end{equation}
We define the interaction context as $\mathcal{T} = \left(o_{0:t}, a_{1:t}\right)$ and 
analyze its information content under two mild assumptions: (A1) partial 
observability, $H\left(s_k \mid o_k\right) > 0$, a single image cannot uniquely identify 
the viewpoint or kinematics; and (A2) information-preserving transitions, 
$I\left(s_0; s_k \mid a_{1:k}\right) > 0$, state transitions preserve $\psi$, 
which holds since $\psi$ is time-invariant.

\textbf{Proposition 1.} \textit{Under A1 and A2, for any action sequence $a_{1:t}$, the interaction context 
$\mathcal{T}$ carries strictly more information about $\psi$ than any single 
observation:}
\begin{equation}
I\left(\psi;\, o_{0:t},\, a_{1:t}\right) > I\left(\psi;\, o_0\right).
\end{equation}

The proof is provided in \autoref{appendix:proof}. 
Since the result holds for any action distribution, 
task-agnostic random movements can also enrich the information 
available about $\psi$ despite carrying no task-specific information.

\begin{figure*}
    \centering
    \includegraphics[width=1\linewidth]{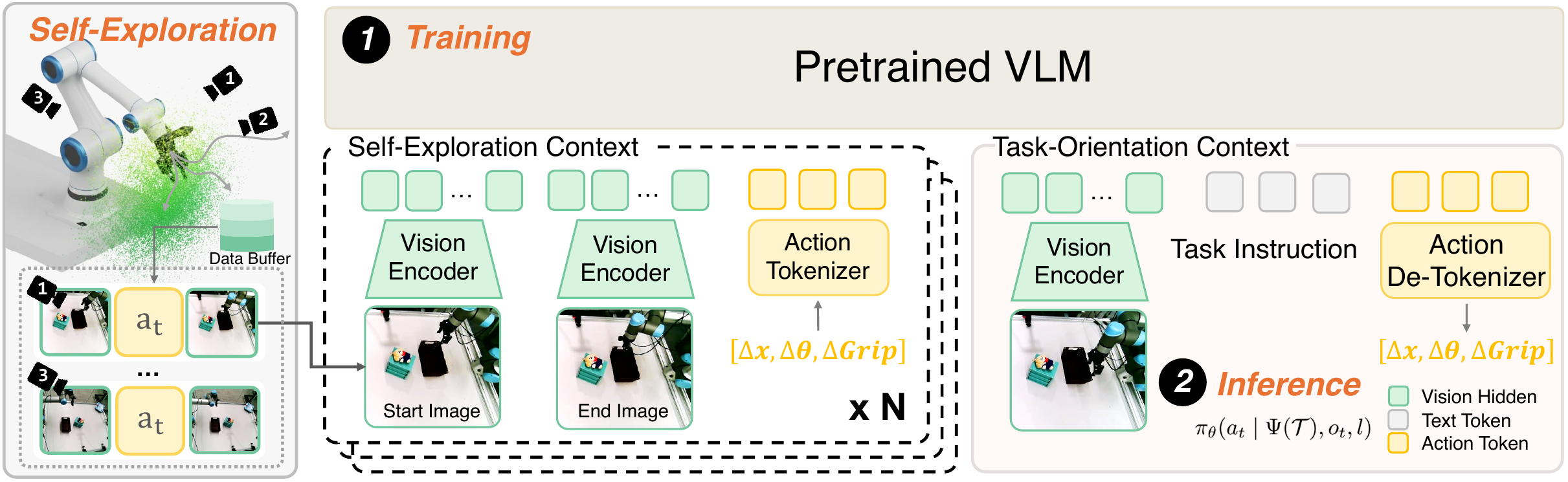}
    \caption{Overview of the ICWM Training and Inference Pipeline. (1) Training: The model is trained on data collected across diverse system configurations, where task-agnostic interaction clips are prepended to each training sample as context. (2) Inference: At test time, the robot first performs stochastic exploration to collect system context, which then guides the policy to generate precise actions via in-context inference.}
    \label{fig:methods}
\end{figure*}

\section{In-Context World Modeling}
\label{sec:method}

To resolve the information incompleteness identified in \autoref{sec:analysis}, we introduce In-Context World Modeling (ICWM) (\autoref{fig:methods}). The core philosophy of ICWM is to transform the VLA from a static mapping into an adaptive inference mechanism that recovers the latent configuration $\psi$ through environmental interactions. 

Specifically, given a short history of task-agnostic interaction clips $\mathcal{T} = \left\{\left(o^s_i, a_i, o^e_i\right) \right\}_{i=1}^{N}$, the configuration can be inferred implicitly with a function $\Psi\left(\mathcal{T}\right)$. By conditioning the policy on interactions collected under the current configuration $\psi$, the policy is reformulated as:
\begin{equation}
a_t \sim \pi_\theta\left(a_t \mid \Psi \left(\mathcal{T}\right), o_t, l\right).
\end{equation}

The $\Psi\left(\mathcal{T}\right)$ term serves as a representation of the world dynamics under $\psi$, enabling the model to interpret $o_t$ in the context. This interaction-centric conditioning allows the policy to adapt its action selection to the specific visual and physical setup encountered at test time, thereby overcoming the generalization collapse of standard VLA models.

\subsection{In-Context Training}
\label{subsec:training}

To implement $\Psi\left(\mathcal{T}\right)$, we adopt a parameter-efficient design where $\Psi$ 
shares its parameters with the VLA backbone $\pi_\theta$, motivated by the structural symmetry between action prediction and configuration inference—both require understanding the correspondence between observations and actions. 

Each training sample is constructed by prepending $N$ task-agnostic interaction clips to the task query. 
These clips are randomly sampled from a pool of interaction segments collected across all training trajectories and viewpoints, ensuring diversity in the interaction context.
The model is trained on data collected across diverse system configurations, where the interaction context $\mathcal{T}$ naturally varies with the 
underlying $\psi$. This variation provides an implicit training signal: to accurately predict task actions across configurations, the model must learn to extract and utilize 
the dynamics information carried in $\mathcal{T}$, implicitly modeling the action-to-observation mapping under each specific system setup. The model is trained with 
the following loss:
\begin{equation}
\mathcal{L} = -\log \pi_\theta\left(a_t \mid \Psi\left(\mathcal{T}\right), o_t, l\right),
\end{equation}
where $\Psi\left(\mathcal{T}\right)$ denotes the hidden states induced by the 
interaction context.

\subsection{Test-Time Active Probing and Inference}

At deployment, ICWM enables task-specific demonstration-free adaptation to novel system configurations through a two-phase inference protocol that requires no gradient updates, prior calibration, or knowledge of the target environment.

\textbf{Active Probing Phase.} 
Before task execution, the robot collects the interaction context $\mathcal{T} = \{\left(o^s_i, a_i, o^e_i\right)\}_{i=1}^N$ by performing $N$ task-agnostic probing actions. For each step $i \in \{1, \dots, N\}$, a random target pose is sampled within the robot's safe workspace, and the robot executes an action $a_i$ to reach it. The resulting transition $\left(o^s_i, a_i, o^e_i\right)$ is recorded. These probing movements are designed to be spatially diverse, sampling multiple directions relative to the end-effector's pose to provide sufficient coverage of the local dynamics manifold.
The probing workspace is defined to avoid contact with task-relevant objects, ensuring that the task initial state remains undisturbed throughout this phase.

\textbf{In-Context Execution Phase.} Once $\mathcal{T}$ is collected, the 
configuration inference function $\Psi$ processes the interaction context to implicitly recover the 
latent system configuration. Conditioned on the inferred hidden representation $\Psi\left(\mathcal{T}\right)$, the current observation $o_t$, and the language instruction $l$, the policy generates the task action $a_t \sim \pi_\theta\left(a_t \mid o_t, l, \Psi\left(\mathcal{T}\right)\right)$.
Since $\Psi$ shares parameters with the VLA backbone, this is implemented as a single forward pass where the Transformer first attends to $\mathcal{T}$, building configuration-aware hidden states, before processing the task query $\left(o_t, l\right)$ to produce actions aligned with the deployed physical setup.

\section{Experiments}

In this section, we evaluate the effectiveness of In-Context World Modeling (ICWM) in enabling vision-language-action models to adapt to novel system configurations without parameter updates.
We focus on answering three questions: 

(1) \textit{Generalization}: Does ICWM improve generalization to unseen viewpoints? 

(2) \textit{System Identification}: Does the model truly perform implicit world modeling rather than pattern matching? 

(3) \textit{Versatility}: Does ICWM generalize beyond geometric shifts to semantic and morphological perturbations?

\subsection{Experimental Setup}

\textbf{Simulated Benchmark.} 
We evaluate on LIBERO~\cite{liu2023libero}, 
comprising four task suites (Spatial, Object, Goal, Long) that assess spatial 
reasoning, object understanding, goal conditioning, and long-horizon execution. 
We adopt a cross-view protocol: training on 8 azimuthal angles 
($\psi \in \{30^\circ, 60^\circ, \dots, 330^\circ\}$) and evaluating on both 
in-domain and 6 unseen OOD viewpoints 
($\psi' \in \{45^\circ, 135^\circ, \dots, 315^\circ\}$), yielding 
$500 \times 15 \times 4$ total episodes. Details are in \autoref{app:sim}.

\textbf{Real-Robot Setup.} 
Our platform is a UR5e manipulator with a 
12-camera multi-view system, split into 6 training and 6 held-out test 
viewpoints. We evaluate four manipulation tasks (stacking, lifting, 
pick-and-place) with 25 trials per task per novel viewpoint (600 total), 
reporting the average success rate. Details are in \autoref{app:real}.

\begin{figure}
    \centering
    \includegraphics[width=1\linewidth]{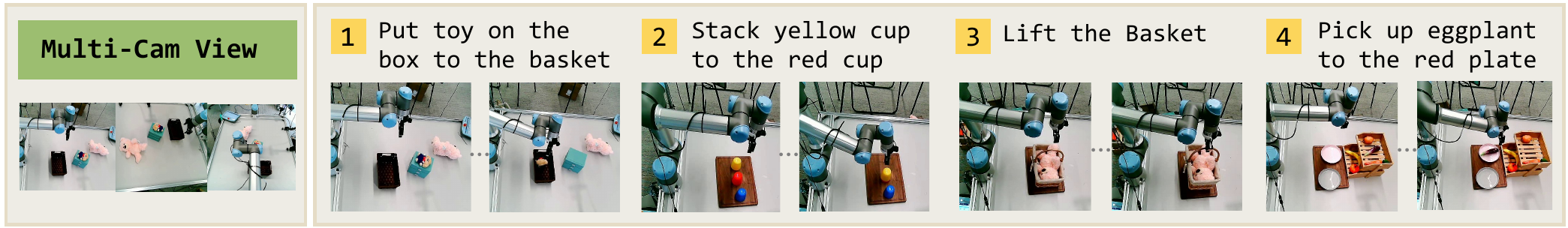}
    \vspace{-5mm}
    \caption{Real-World Task Suite with four manipulation tasks, including stacking, lifting, and pick-and-place, evaluated across six distinct camera viewpoints.}
    \label{fig:placeholder}
    \vspace{-4pt}
\end{figure}

\textbf{Baselines.} 
We evaluate two categories of models:
\textbf{(i) Controlled Comparisons:} (a) \textbf{Multi-View BC (MV)} shares identical architecture and multi-view data with ICWM but lacks context; (b) \textbf{Explicit Configuration (EXP)} augments MV with ground-truth camera angles as text inputs. This category isolates the exact contribution of our interaction context.
\textbf{(ii) Contextual References:} Pretrained models, including NORA~\cite{Hung2025NORAAS},$\pi$-FAST~\cite{fastplus}, and $\pi_{0.5}$~\cite{Intelligence202505AV}, are fine-tuned on a single viewpoint. They serve not as direct benchmarks, but to demonstrate that out-of-the-box generalization remains an unsolved challenge even under large-scale pretraining.

\textbf{Implementation Details.}
Our model utilizes Qwen2.5-VL-3B~\cite{Bai2025Qwen25VLTR} as the backbone and FAST~\cite{fastplus} as the action tokenizer, with action chunk size of 5 and $N{=}5$ context clips. Implemented in PyTorch and trained on 8 NVIDIA A100 GPUs, the system is optimized via AdamW with a weight decay of $10^{-4}$ and a learning rate peaking at $5 \times 10^{-5}$ after a 50k-step warmup, followed by cosine decay. The implementation details of self-exploration is detailed in \autoref{app:random}.

\begin{figure*}
    \centering
    \includegraphics[width=1\linewidth]{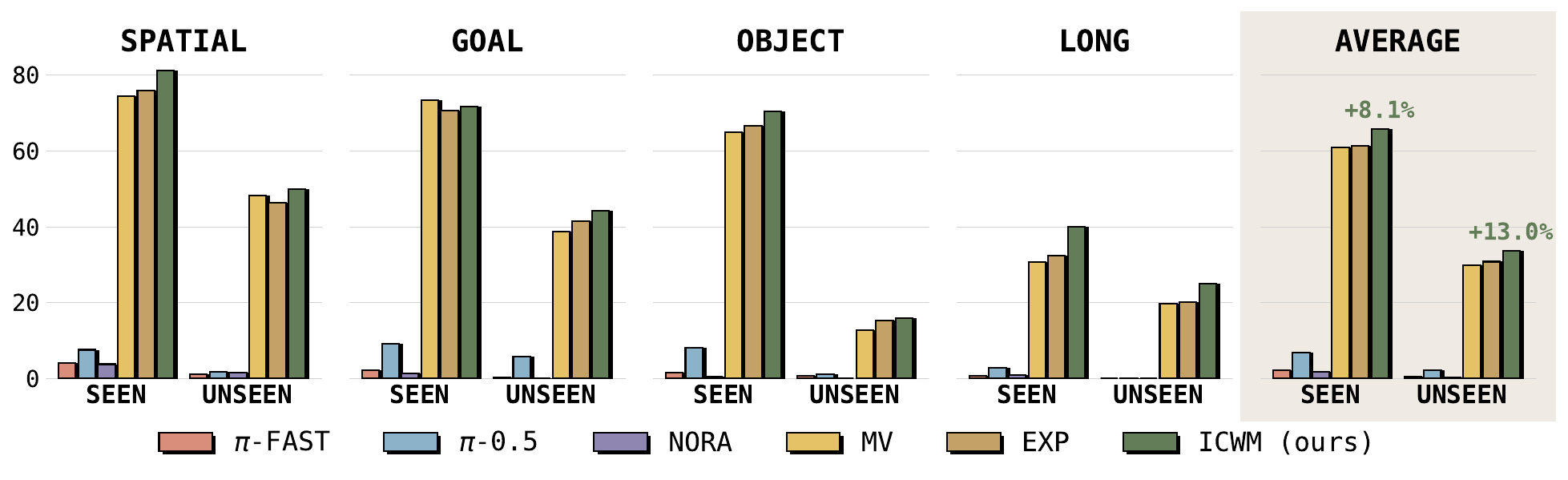}
    \vspace{-7mm}
    \caption{Success Rates (\%) on LIBERO for Seen (In-Domain) and Unseen (OOD) Viewpoints.}
    \label{fig:simulation}
\end{figure*}

\subsection{Simulation Results}

As shown in \autoref{fig:simulation}, ICWM consistently outperforms all baselines in both seen and unseen viewpoints, leading to several key observations:
\textbf{(1) Viewpoint generalization is a fundamental challenge.} 
While a seen-to-unseen performance drop is observed globally, ICWM demonstrates significantly stronger resilience, improving the OOD success rate by 13.0\% over the Multi-View BC baseline. This confirms that while multi-view training expands spatial data, geometric extrapolation remains an open challenge that standard imitation learning struggles to resolve without test-time adaptation.
\textbf{(2) Implicit identification outperforms explicit specification.} 
ICWM improves OOD success rate by 9.5\% over Explicit Configuration.  
Even with ground-truth camera angles, Exp lacks OOD generalization and an understanding of system dynamics, strengths that ICWM gains from interaction context.
\textbf{(3) ICWM yields the greatest gains on long-horizon tasks.} 
On LIBERO-Long, ICWM surpasses MV by 29.9\% (seen) and 
26.3\% (unseen), the largest relative margins across all 
suites. This is because long-horizon tasks amplify small spatial errors from viewpoint shift, causing cascading failures in baselines, while ICWM mitigates this by continuously grounding actions in system dynamics.

\subsection{Real-world Results}

\begin{wrapfigure}{r}{0.5\textwidth}
    \centering
    \vspace{-7mm}
    \includegraphics[width=\linewidth]{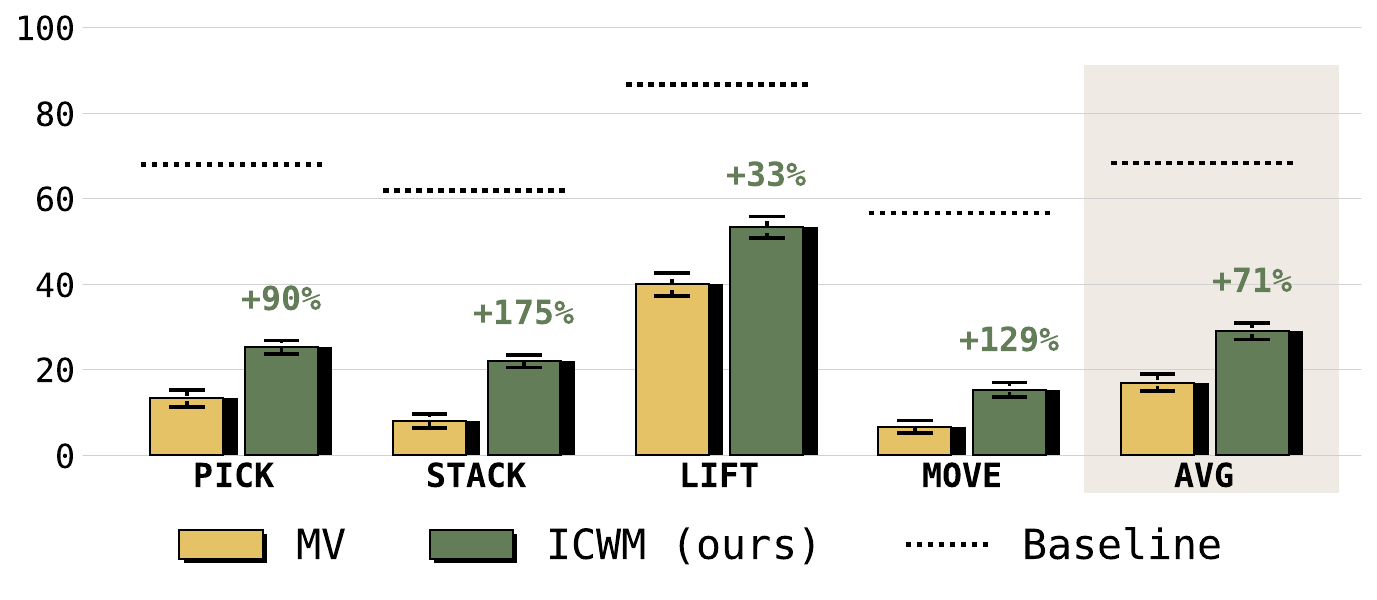}
    \vspace{-5mm}
    \caption{Real-world evaluation on the UR5e platform. ICWM substantially outperforms the Multi-View (MV) baseline under novel viewpoints.}
    \label{fig:real}
    \vspace{-3mm}
\end{wrapfigure}

\begin{figure*}[t]
    \centering
    \includegraphics[width=1\linewidth]{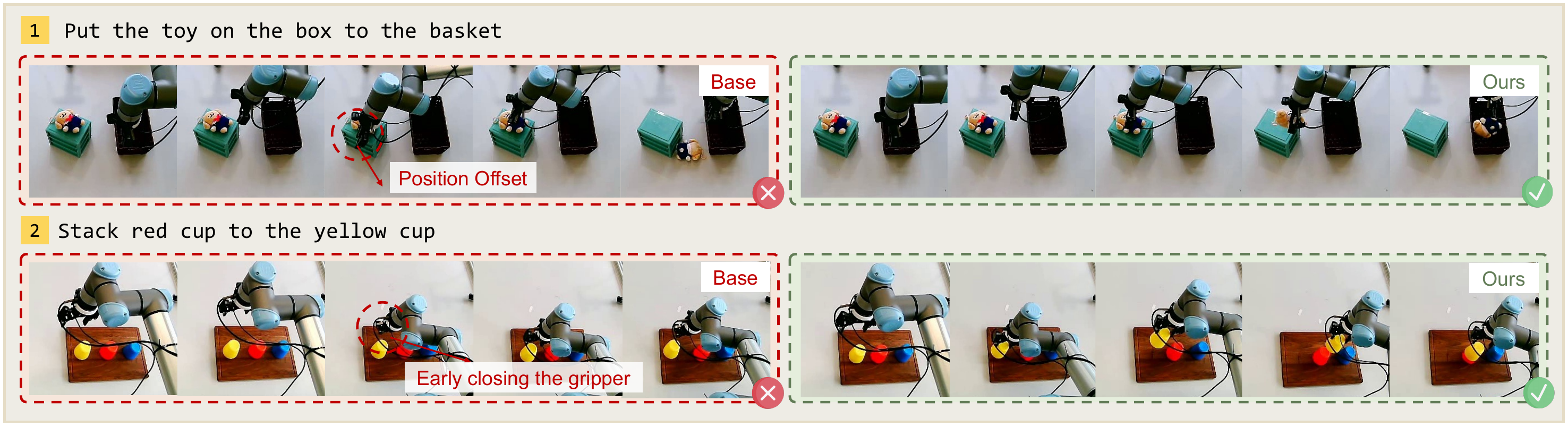}
    \caption{Qualitative Comparison. Without ICWM, standard policies exhibit (a) position offsets or (b) premature gripper closure due to viewpoint shifts.}
    \label{fig:case}
    \vspace{-4pt}
\end{figure*}

As shown in \autoref{fig:real}, standard VLA performance drops sharply from 68\% to 17\% upon viewpoint shift, confirming that standard mix-training alone cannot bridge the gap to novel configurations at test time. In contrast, by introducing In-Context World Modeling (ICWM), the policy effectively mitigates this degradation without any parameter updates or task-specific demonstrations. As qualitatively shown in \autoref{fig:case}, while the base VLA exhibits end-effector drift and repeated grasp failures due to spatial misalignment, our ICWM enhanced robot utilizes the context window as a dynamic calibration frame, achieving precise manipulation by grounding its actions in the current environment's specific geometry.

\section{Analysis}

\subsection{What Does the Interaction Context Contribute?}

\begin{figure}[t]
    \centering
    \begin{minipage}{0.52\linewidth}
        \centering
        \vspace{-3pt}
        \captionof{table}{Ablation on Interaction Context. Removing any context component degrades performance, confirming that joint observation of actions and outcomes is essential for system dynamics.}
        \label{tab:ablation}
        \resizebox{\linewidth}{!}{
        \setlength{\tabcolsep}{4pt}
        \begin{tabular}{lccccc}\toprule
&ICWM &w/o act. &w/o img. &w/o ctx. & false ctx. \\
\midrule
$45^\circ$  & \cellcolor{eggshell}36.6 & 30.8{\tiny\textcolor[HTML]{006400}{$\downarrow$\,15.8\%}} & 17.6{\tiny\textcolor[HTML]{006400}{$\downarrow$\,51.9\%}} & 32.4{\tiny\textcolor[HTML]{006400}{$\downarrow$\,11.5\%}} & 27.8{\tiny\textcolor[HTML]{006400}{$\downarrow$\,24.0\%}} \\
$135^\circ$ & \cellcolor{eggshell}2.2  & 2.0{\tiny\textcolor[HTML]{006400}{$\downarrow$\,9.1\%}}  & 0.8{\tiny\textcolor[HTML]{006400}{$\downarrow$\,63.6\%}} & 1.6{\tiny\textcolor[HTML]{006400}{$\downarrow$\,27.3\%}}  & 1.2{\tiny\textcolor[HTML]{006400}{$\downarrow$\,45.5\%}} \\
$225^\circ$ & \cellcolor{eggshell}8.8  & 8.2{\tiny\textcolor[HTML]{006400}{$\downarrow$\,6.8\%}}  & 3.0{\tiny\textcolor[HTML]{006400}{$\downarrow$\,65.9\%}} & 7.8{\tiny\textcolor[HTML]{006400}{$\downarrow$\,11.4\%}}  & 4.2{\tiny\textcolor[HTML]{006400}{$\downarrow$\,52.3\%}} \\
$255^\circ$ & \cellcolor{eggshell}28.4 & 28.0{\tiny\textcolor[HTML]{006400}{$\downarrow$\,1.4\%}}  & 13.2{\tiny\textcolor[HTML]{006400}{$\downarrow$\,53.5\%}}& 27.8{\tiny\textcolor[HTML]{006400}{$\downarrow$\,2.1\%}}  & 24.4{\tiny\textcolor[HTML]{006400}{$\downarrow$\,14.1\%}} \\
$285^\circ$ & \cellcolor{eggshell}36.6 & 25.6{\tiny\textcolor[HTML]{006400}{$\downarrow$\,30.1\%}} & 14.6{\tiny\textcolor[HTML]{006400}{$\downarrow$\,60.1\%}}& 27.4{\tiny\textcolor[HTML]{006400}{$\downarrow$\,25.1\%}} & 23.6{\tiny\textcolor[HTML]{006400}{$\downarrow$\,35.5\%}} \\
$315^\circ$ & \cellcolor{eggshell}37.6 & 35.0{\tiny\textcolor[HTML]{006400}{$\downarrow$\,6.9\%}}  & 16.2{\tiny\textcolor[HTML]{006400}{$\downarrow$\,56.9\%}}& 35.0{\tiny\textcolor[HTML]{006400}{$\downarrow$\,6.9\%}}  & 32.4{\tiny\textcolor[HTML]{006400}{$\downarrow$\,13.8\%}} \\
\midrule
avg & \cellcolor{eggshell}25.0 & 21.6{\tiny\textcolor[HTML]{006400}{$\downarrow$\,13.6\%}} & 10.9{\tiny\textcolor[HTML]{006400}{$\downarrow$\,56.4\%}} & 22.0{\tiny\textcolor[HTML]{006400}{$\downarrow$\,12.0\%}} & 18.9{\tiny\textcolor[HTML]{006400}{$\downarrow$\,24.4\%}} \\
\bottomrule
\end{tabular}
        }
    \end{minipage}
    \hfill
    \begin{minipage}{0.46\linewidth}
        \centering
        \includegraphics[width=\linewidth]{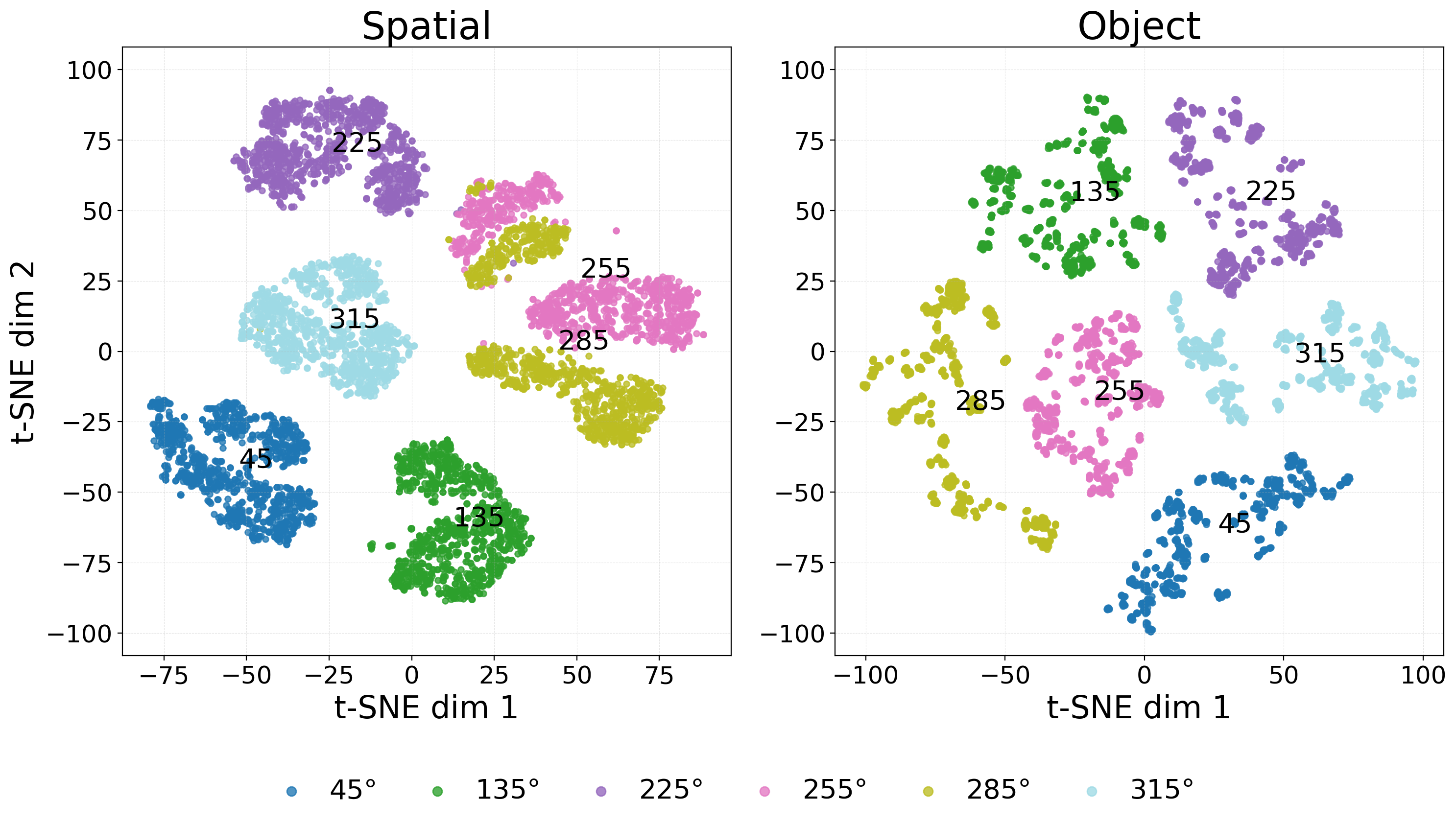}
        \vspace{-3pt}
        \captionof{figure}{t-SNE of $\Psi\left(\mathcal{T}\right)$ across OOD viewpoints (perplexity=30).}
        \label{fig:tsne}
    \end{minipage}
\end{figure}

\textbf{Ablation on context components.} 
We ablate interaction context across five settings: (1) full-context (ICWM), (2) \textit{w/o actions} (action tokens omitted), (3) \textit{w/o images} (image tokens omitted), (4) \textit{w/o context} (all interaction tokens excluded), and (5) \textit{false context} (clips from a $180^\circ$-offset viewpoint).
The asymmetric drops in Tab.~\ref{tab:ablation} reveal several key findings: \textit{w/o images} causes the largest collapse (avg. $-56.4\%$)---without visual outcomes, the model mimics exploratory actions as task demonstrations, actively hurting performance; \textit{w/o actions} degrades more moderately, confirming that visual flow provides a coarse spatial anchor but full calibration requires the paired $(o^s_i, a_i, o^e_i)$ tuple.
Critically, \textit{false context} performs worse than no context at all ($18.9$ vs.\ $22.0$), indicating that misaligned context actively misleads the policy's world model rather than being passively ignored. This negative transfer, symmetric in magnitude to the gains from correct context ($+13.6\%$), confirms that the model genuinely conditions on context content for configuration inference.

\textbf{Is in-context capability an emergent property of specialized training?} 
To determine whether in-context adaptation arises from standard sequence modeling, we evaluate a BC policy trained without in-context supervision under the same interaction context. Performance collapses to near-zero ($<1\%$) when interaction tokens are prepended, confirming that this capability must be explicitly incentivized during training rather than emerging naturally from imitation learning.

\textbf{Are the Learned Implicit Representations Identifiable?} 
We examine whether $\Psi(\mathcal{T})$ forms a structured representation with respect to system configuration.
We visualize the hidden representations $\Psi(\mathcal{T})$ via t-SNE across the six OOD viewpoints (1,024 points per viewpoint). 
As shown in Fig.~\ref{fig:tsne}, the representations exhibit tight within-viewpoint clustering, demonstrating stability, and clear between-viewpoint separation, demonstrating identifiability.

\subsection{What Probing Strategy is Effective?}
\label{subsec:probing}

\begin{wraptable}{r}{0.5\textwidth}
\centering
\small
\vspace{-5mm}
\caption{Success rates (\%) under different probing strategies on OOD viewpoints.}
\label{tab:probing}
\resizebox{0.5\textwidth}{!}{ 
\begin{tabular}{lccccc}
\toprule
 & MV & Random & R-only & Z-only & XY-only \\
\midrule
45°  & \cellcolor{gray!20} 30.4 & \textbf{36.6} & 35.8 & 34.6 & 34.6 \\
135° & \cellcolor{gray!20} 1.2  & 2.2  & 1.0  & \textbf{3.0}  & 1.0  \\
225° & \cellcolor{gray!20} 2.8  & 8.8  & 6.0  & 13.0 & \textbf{14.8} \\
255° & \cellcolor{gray!20} 24.0 & 28.4 & 30.4 & 25.2 & \textbf{30.6} \\
285° & \cellcolor{gray!20} 32.8 & 36.6 & \textbf{38.8} & 31.2 & 33.6 \\
315° & \cellcolor{gray!20} 27.6 & \textbf{37.6} & 29.2 & 30.0 & 35.0 \\
\midrule
Avg  & \cellcolor{gray!20} 19.8 & \textbf{25.0} & 23.4 & 22.8 & 24.9 \\
\bottomrule
\end{tabular}
}
\vspace{-10pt}
\end{wraptable}
A practical question for deploying ICWM is whether the choice of probing strategy matters. 
We evaluate four strategies: (a) \textit{Random}, which samples target poses uniformly across all spatial directions; 
(b) \textit{XY-only}, which restricts movements to the horizontal plane; (c)
\textit{Z-only}, which moves exclusively along the vertical axis; and (d) \textit{R-only}, which varies only the end-effector orientation.
As shown in \autoref{tab:probing}, all four strategies consistently outperform the multi-view baseline by 15–27\%, confirming that the benefit of ICWM stems from the interaction format itself rather than any particular movement pattern. 
Performance differences across strategies suggest that different axes expose different aspects of the local dynamics manifold, with no single strategy dominating across all viewpoints.

\subsection{Does ICWM Generalize Beyond Camera Shifts?}

To evaluate whether ICWM generalizes to different deployment settings, we stress-test the same trained model against two distinct categories of out-of-distribution perturbation: semantic scene variations and physical morphological changes. Experimental details are provided in \autoref{app:gen}.

\begin{figure}
    \centering
    \begin{subfigure}[b]{0.595\linewidth}
        \centering
        \includegraphics[width=\linewidth]{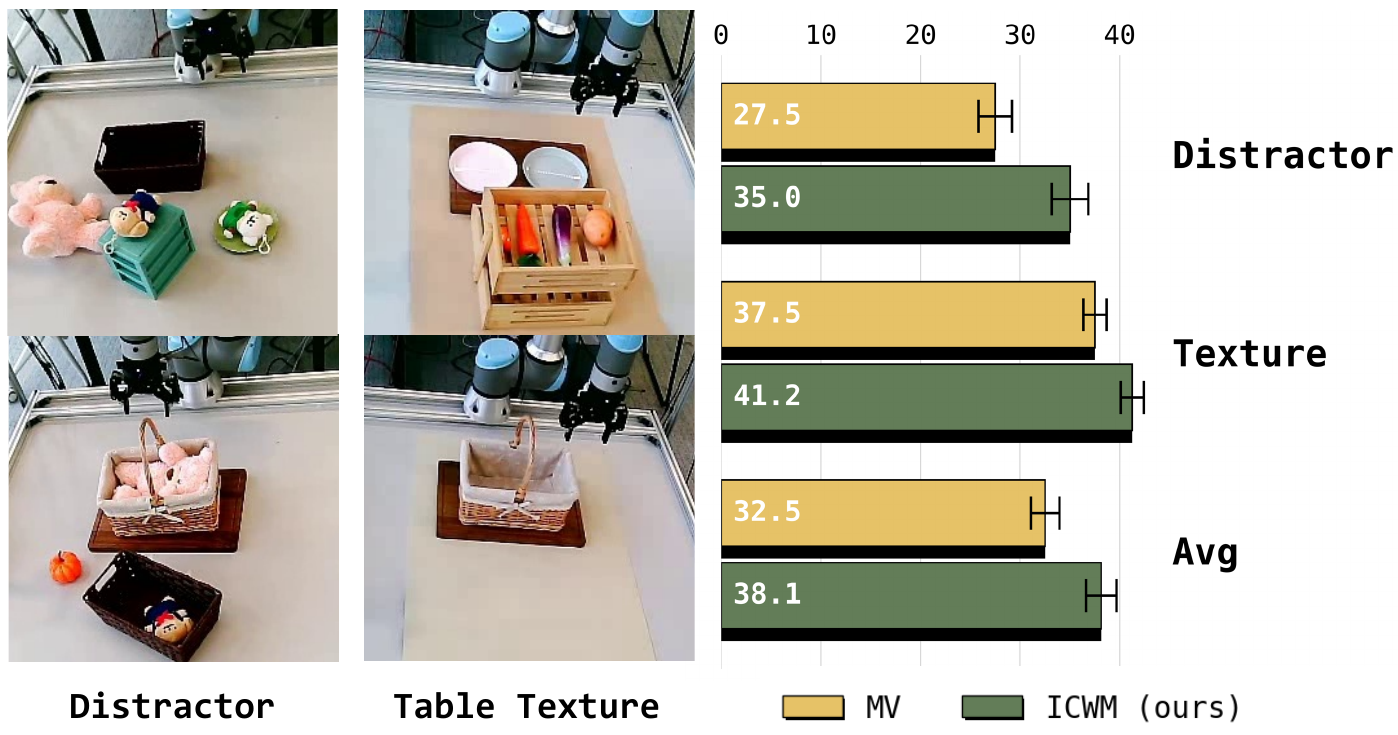}
        \caption{Semantic perturbations}
        \label{fig:gen}
    \end{subfigure}
    \hfill
    \begin{subfigure}[b]{0.395\linewidth}
        \centering
        \includegraphics[width=\linewidth]{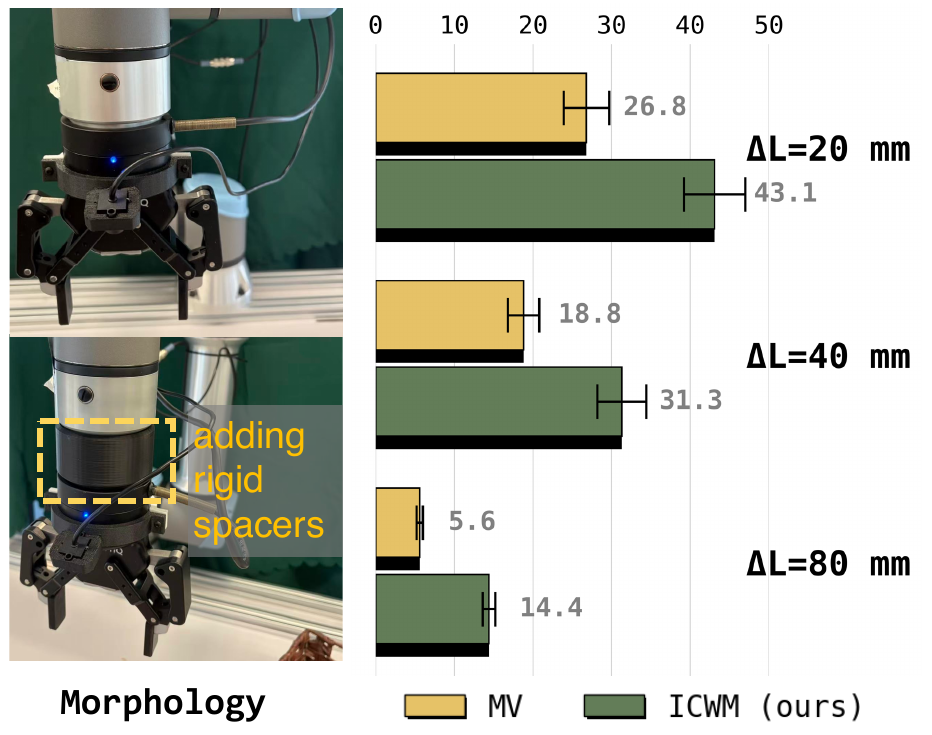}
        \caption{Morphological generalization}
        \label{fig:morph}
    \end{subfigure}
    \vspace{-4pt}
    \caption{Robustness to semantic scene variations and robot morphological changes. ICWM maintains a performance margin over the baseline when facing novel out-of-distribution perturbations.}
\end{figure}

\noindent\textbf{Semantic perturbations.} 
As shown in (\autoref{fig:gen}), ICWM maintains a consistent margin over MV under both distractor objects (35.0 vs. 27.5) and novel table textures (41.2 vs. 37.5).
The more moderate gains relative to viewpoint generalization likely reflect the scarcity of diverse scene-configuration data in current datasets rather than a fundamental limitation of the mechanism.

\begin{wrapfigure}{r}{0.36\textwidth}
    \vspace{-8mm}
    \centering
    \includegraphics[width=\linewidth]{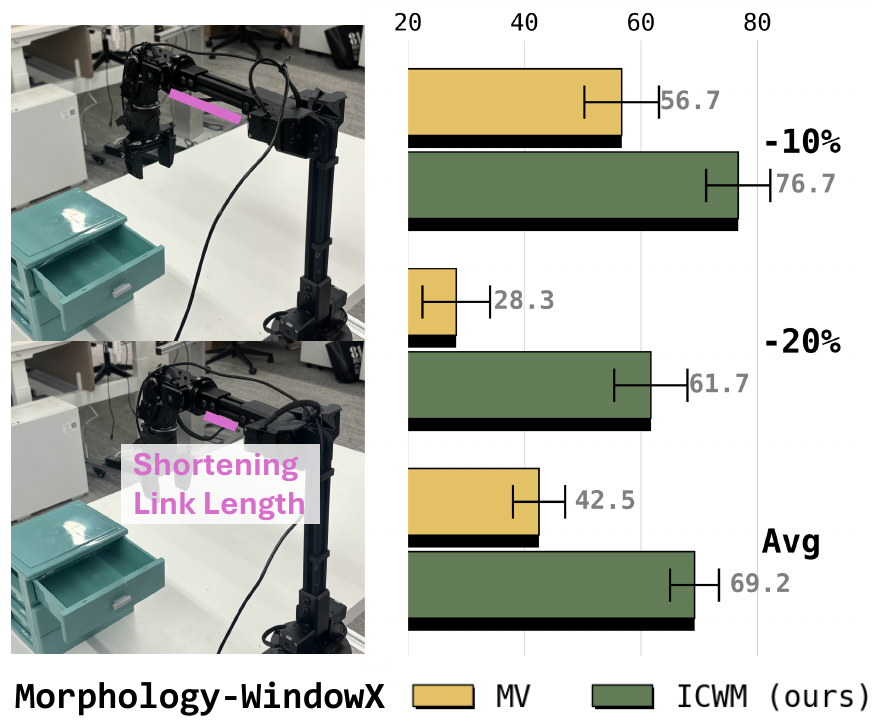}
    \vspace{-7mm}
    \caption{Morphological generalization on WindowX.}
    \label{fig:gen_windowx}
    \vspace{-10mm}
\end{wrapfigure}

\noindent\textbf{Morphological generalization.} Attaching rigid spacers ($\Delta L \in \{20, 40, 80\}$ mm) to the gripper flange alters forward kinematics at test time. ICWM consistently outperforms MV across all offsets with a stable margin of +60\% (\autoref{fig:gen_windowx}); at $\Delta L = 80$ mm, where MV largely fails (5.6), ICWM retains non-trivial success (14.4) by inferring the altered kinematics from probing context.

We further validate this on a WindowX platform with systematically varied link lengths (\{100\%, 90\%, 80\%, 70\%\} of the original), training on two boundary configurations (100\% and 70\% link length) and evaluating zero-shot on two interpolated OOD configurations (90\% and 80\%). 
As the link-length offset increases from 10\% to 20\%, MV's average success rate collapses by more than half (57\% $\to$ 28\%), while ICWM degrades far more gracefully (77\% $\to$ 62\%), widening its margin over MV from 20 to 34 points (\autoref{fig:gen_windowx}). This mirrors the pattern observed with spacer attachments: as kinematic uncertainty grows, ICWM's advantage over MV grows with it, consistent with the benefit stemming from explicit kinematic inference via the probing context.

\subsection{Inference Latency Analysis}

\begin{wrapfigure}{r}{0.5\textwidth}
    \vspace{-5mm}
    \centering
    \includegraphics[width=\linewidth]{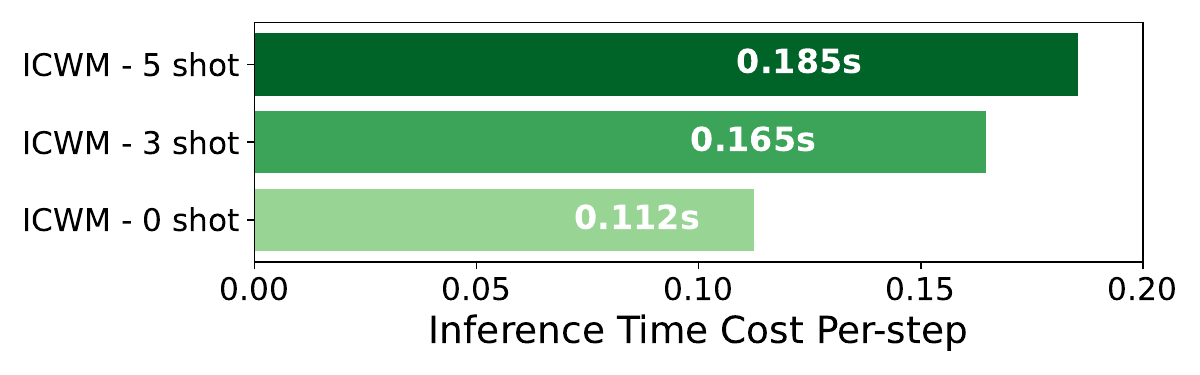}
    \vspace{-5mm}
    \caption{Comparison of inference time across different setting.}
    \label{fig:time}
    \vspace{-5mm}
\end{wrapfigure}

We evaluate the computational overhead of ICWM on a single NVIDIA RTX 4090 GPU. The extra latency is mainly due to the $2N$ image and $N$ action tokens. While the baseline VLA requires 0.112s per inference step, our ICWM-enhanced model with $N=3$ and $N=5$ context clips incurs a latency of 0.165s and 0.185s respectively, without compromising control loop stability (\autoref{fig:time}). 
Furthermore, since the interaction context $\tilde{\mathcal{T}}_\psi$ is static under a fixed configuration, its hidden states can be pre-computed and reused via KV caching, effectively reducing per-step inference cost back to near-baseline levels.

\section{Conclusion}

In this work, we identified the absence of explicit system configuration conditioning in modern VLA models as an underexplored limitation to generalization. 
Inspired by human motor adaptation, we proposed In-Context World Modeling (ICWM), which transforms the VLA policy from a static mapping into an adaptive inference engine. 
By conditioning the policy on self-generated forward interaction clips, ICWM implicitly captures the underlying sensory-motor relationship at test time without any parameter updates.
Experiments across simulation benchmarks and real-world platforms demonstrate that ICWM substantially reduces spatial ambiguities under novel viewpoints, and can also extend to semantic scene variations and robot morphological changes.

\clearpage
\bibliographystyle{unsrtnat} 
\bibliography{main}

\clearpage
\beginappendix


\section{Proof of Proposition 1}
\label{appendix:proof}

We prove that the interaction context $\mathcal{T} = \left(o_{0:t}, a_{1:t}\right)$ 
carries strictly more information about the system configuration $\psi$ than 
a single observation $o_0$ alone.

\begin{proof}
Since $\psi \subseteq s_0$, it suffices to prove the stronger statement 
$I\left(s_0; o_{0:t} \mid a_{1:t}\right) > I\left(s_0; o_0\right)$, from which the theorem 
follows by the data processing inequality.

Applying the chain rule of mutual information:
\begin{equation}
I\left(s_0;\, o_{0:t} \mid a_{1:t}\right) = I\left(s_0;\, o_0 \mid a_{1:t}\right) + 
I\left(s_0;\, o_{1:t} \mid o_0, a_{1:t}\right).
\label{eq:chain}
\end{equation}

\textit{First term.} In the graphical model, $s_0$ is a root node and 
$a_{1:t}$ are exogenous root nodes. Every path between $s_0$ (or $o_0$) 
and any $a_k$ passes through a collider at $s_k$ via $s_{k-1} \to s_k 
\leftarrow a_k$. Since no collider or its descendants are conditioned upon, 
these paths are blocked by d-separation, giving $\left(s_0, o_0\right) \perp a_{1:t}$. 
Therefore:
\begin{equation}
I\left(s_0;\, o_0 \mid a_{1:t}\right) = I\left(s_0;\, o_0\right).
\label{eq:first}
\end{equation}

\textit{Second term.} We show $I\left(s_0;\, o_{1:t} \mid o_0, a_{1:t}\right) > 0$. 
Consider the path $s_0 \to s_1 \to \cdots \to s_k \to o_k$ for any 
$k \geq 1$. Conditioning on $o_0$—a descendant of $s_0$—activates the 
collider at $s_0$, leaving the path $s_0 \to s_k \to o_k$ active under 
d-separation given $\{o_0, a_{1:t}\}$. Hence $s_0 \not\perp o_k \mid 
o_0, a_{1:t}$. By A2, the state chain preserves information about $s_0$, 
so:
\begin{equation}
I\left(s_0;\, o_{1:t} \mid o_0, a_{1:t}\right) \geq I\left(s_0;\, o_k \mid o_0, a_{1:t}\right) 
> 0.
\label{eq:second}
\end{equation}

Substituting \eqref{eq:first} and \eqref{eq:second} into \eqref{eq:chain}:
\begin{equation}
I\left(s_0;\, o_{0:t} \mid a_{1:t}\right) = I\left(s_0;\, o_0\right) + 
\underbrace{I\left(s_0;\, o_{1:t} \mid o_0, a_{1:t}\right)}_{>\ 0} > I\left(s_0;\, o_0\right).
\qedhere
\end{equation}
\end{proof}

\section{Details of Simulated Experiment Setup}
\label{app:sim}

\subsection{Experimental Configuration}

\begin{figure}
    \centering
    \includegraphics[width=1\linewidth]{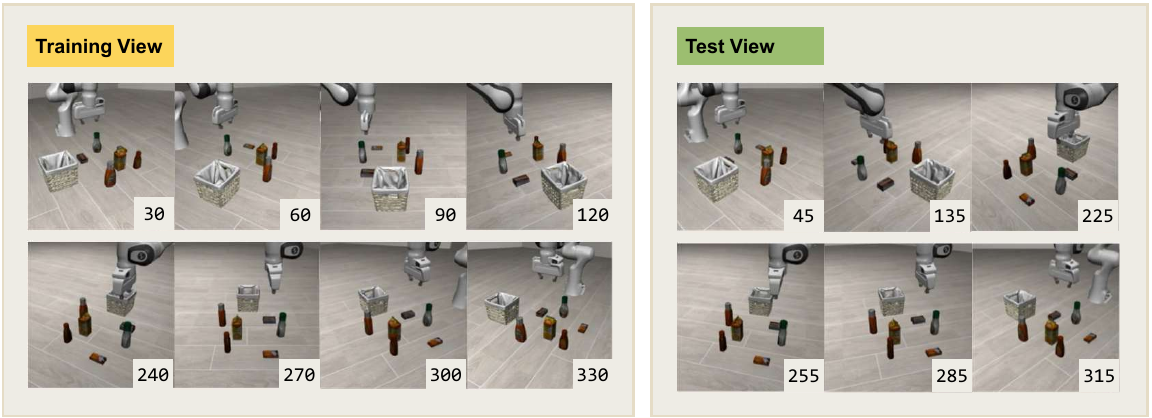}
    \caption{Viewpoint Distribution for Training and Evaluation. We illustrate the 8 in-domain training angles and 6 out-of-domain testing angles used to benchmark viewpoint generalization.}
    \label{fig:viewpoints_demo}
\end{figure}

\paragraph{System Identification Benchmark Design.} Unlike the standard LIBERO setup, which typically evaluates policies under a fixed, single camera viewpoint, our experimental protocol is specifically designed to stress-test OOD viewpoint generalization. We distribute 14 discrete azimuthal angles around the workspace center. We designate 8 In-Domain (ID) angles for training: $\psi_{train} \in \{30^\circ, 60^\circ, 90^\circ, 120^\circ, 240^\circ, 270^\circ, 300^\circ, 330^\circ\}$, while withholding 6 Out-of-Domain (OOD) angles: $\psi_{test} \in \{45^\circ, 135^\circ, 225^\circ, 255^\circ, 285^\circ, 315^\circ\}$ exclusively for evaluation. \autoref{fig:viewpoints_demo} illustrates the drastic visual shifts between these angles, which introduce spatial ambiguities that standard VLA policies struggle to resolve without explicit system identification.

\paragraph{LIBERO Suite Specifications.} 
To evaluate the versatile capabilities of our model, we utilize the LIBERO benchmark \cite{liu2023libero}, categorized into four specialized suites:
\begin{itemize} [itemsep=0.3em, topsep=0.5em]
    \item LIBERO-Spatial: Benchmarks spatial reasoning by presenting identical objects in randomized initial layouts, requiring the agent to disambiguate relative geometric cues (e.g., ``pick up the black bowl next to the plate and place it on the plate'').
    \item LIBERO-Goal: Evaluates goal-conditioning by assigning multiple distinct objectives to the same scene configuration, assessing whether the policy can map different instructions to varied behavioral distributions (e.g., ``open the top drawer and put the bowl inside'').
    \item LIBERO-Object: Focuses on semantic grounding by introducing a diverse set of object categories within fixed layouts, testing the model's ability to generalize to novel visual appearances (e.g., ``pick up the bbq sauce and place it in the basket'').
    \item LIBERO-Long: Challenges the model with 10 complex, multi-stage manipulation sequences that require sustained temporal consistency and the ability to recover from small execution errors over extended horizons (e.g., ``turn on the stove and put the moka pot on it'').
\end{itemize}

\paragraph{Data Synthesis and Preprocessing.} 
The training data is generated by replaying expert demonstration trajectories from the original benchmark in the simulator and re-rendering them from the 8 ID camera poses. Following the preprocessing protocol of OpenVLA \cite{openvla}, we filter out unsuccessful episodes and remove redundant frames where action norms are near-zero and the gripper state is static, ensuring a high-density learning signal.

\subsection{Detailed Results}

\autoref{tab: seen} and \autoref{tab: unseen} provide the exhaustive breakdown of success rates for ID and OOD viewpoints. 
Notably, ICWM maintains robust performance across the majority of evaluated conditions, outperforming multi-view behavior cloning and explicit configuration approaches in most settings.
In particular, for LIBERO-Long, ICWM achieves a $25.0\%$ average OOD success rate, a significant margin over the $19.8\%$ achieved by the Multi-View (MV) baseline, demonstrating the power of in-context calibration for complex tasks. 

We also observe that certain OOD viewpoints, particularly 135°, pose challenges for all methods, including ICWM. We attribute this to viewpoint-specific geometric constraints: at 135°, the camera angle may introduce a certain degree of object occlusion and reduce the effective workspace visible to the model, which can cause manipulation targets to occasionally exit the field of view during execution. This suggests a perceptual limitation shared across methods, rather than a failure specific to ICWM alone.

\begin{figure}
    \centering
    \includegraphics[width=0.85\linewidth]{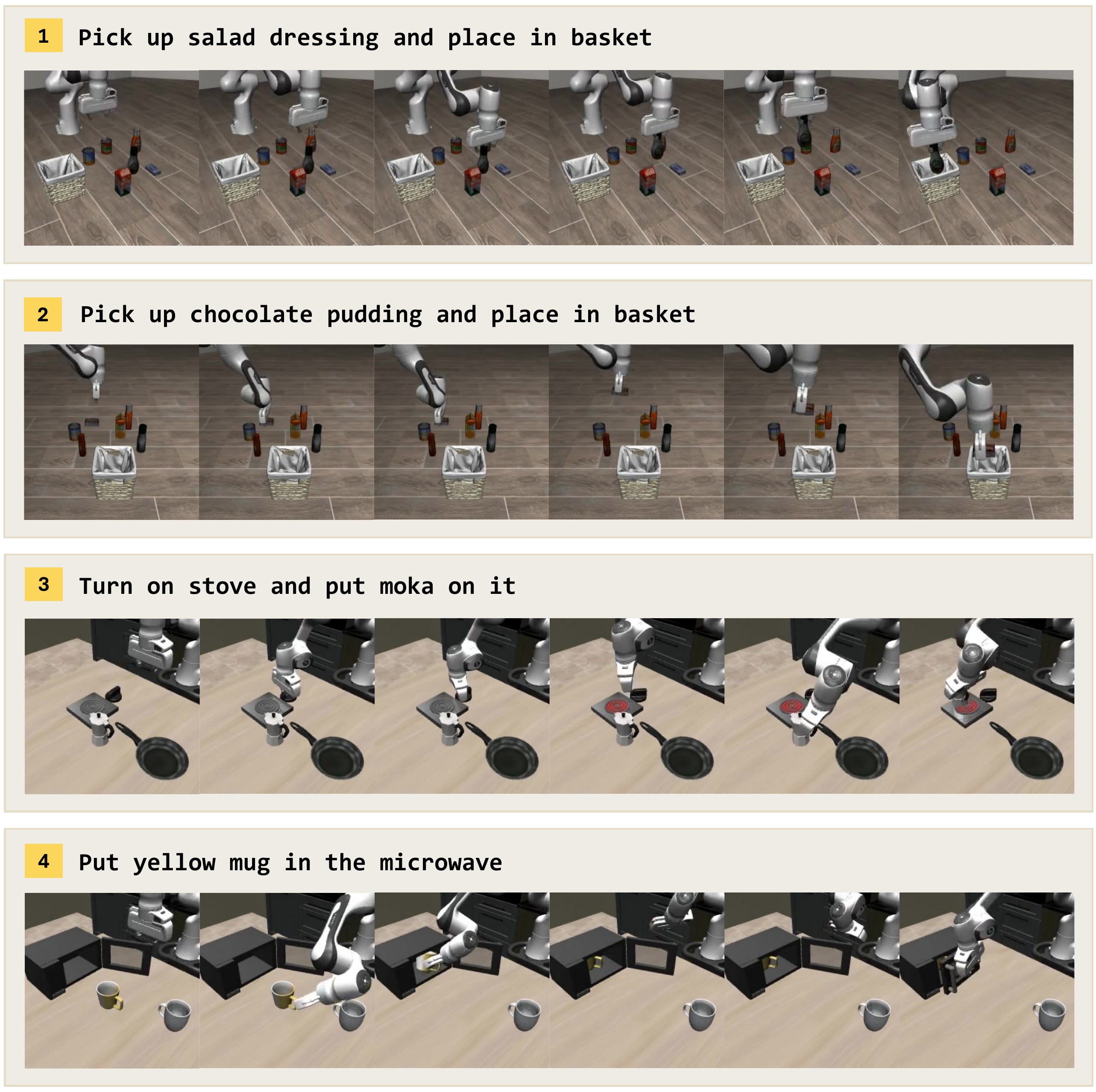}
    \caption{Visualization of Successful Task Rollouts in Simulation. By utilizing the interaction prefix to resolve viewpoint-induced ambiguities, the agent achieves precise grasping and multi-stage execution (e.g., turning on a stove) without any environment-specific fine-tuning.}
    \label{fig:rollouts}
\end{figure}

To further investigate the model's behavior, we visualize the rollout trajectories in Figure~\ref{fig:rollouts}. ICWM demonstrates remarkable execution stability; even when the initial viewpoint causes a spatial offset, the model utilizes the in-context world model to ``re-align'' its end-effector during the first few steps of the task, ensuring a successful grasp that baselines frequently miss due to viewpoint-induced depth and coordinate errors.

\begin{table}[t]\centering
\caption{Success Rates (\%) on In-Domain (Seen) Viewpoints.}
\label{tab: seen}
\begin{tabular}{llccccccccc}\toprule
& &$30^\circ$ & $60^\circ$ &$90^\circ$ &$120^\circ$ &$240^\circ$ &$270^\circ$ &$300^\circ$ &$330^\circ$ &Avg \\
\midrule
\multirow{5}{*}{Spatial} &$\pi$-FAST &19.0 &0.0 &0.0 &0.0 &0.0 &0.0 &0.0 &14.0 &4.1 \\
&$\pi_{0.5}$ &39.0 &0.2 &0.0 &0.0 &0.0 &0.0 &0.6 &20.6 &7.6 \\
&NORA &11.4 &0.6 &0.0 &0.0 &0.0 &0.2 &0.0 &17.8 &3.8 \\
&MV &79.4 &77.2 &69.4 &65.4 &72.8 &71.4 &81.8 &78.4 &74.5 \\
&EXP &79.8 &81.8 &68.2 &69.2 &70.6 &75.6 &79.4 &82.6 &75.9 \\
\rowhighlight
&ICWM (ours) &\textbf{84.6} &\textbf{84.0} &\textbf{75.6} &\textbf{73.8} &\textbf{83.0} &\textbf{77.0} &\textbf{83.6} &\textbf{88.0} &\textbf{81.2} \\
\midrule
\multirow{5}{*}{Goal} &$\pi$-FAST &7.8 &0.0 &0.0 &0.0 &0.0 &0.0 &0.0 &10.0 &2.2 \\
&$\pi_{0.5}$ &40.2 &4.6 &0.0 &0.0 &0.0 &0.0 &0.4 &28.0 &9.2 \\
&NORA &5.6 &0.0 &0.4 &0.0 &0.2 &0.0 &0.0 &4.8 &1.4 \\
&MV &78.4 &\textbf{77.2} &68.4 &\textbf{74.0} &\textbf{72.8} &\textbf{69.0} &71.4 &75.4 &\textbf{73.3} \\
&EXP &78.2 &71.0 &66.4 &68.4 &68.0 &62.8 &\textbf{72.8} &\textbf{77.6} &70.7 \\
\rowhighlight
&ICWM (ours) &\textbf{82.2} &75.4 &\textbf{71.0} &69.0 &62.2 &64.6 &72.2 &76.0 &71.6 \\
\midrule
\multirow{5}{*}{Object} &$\pi$-FAST &0.4 &0.0 &0.0 &0.0 &0.0 &0.0 &0.0 &11.8 &1.5 \\
&$\pi_{0.5}$ &13.4 &0.0 &0.0 &0.0 &0.0 &0.0 &7.6 &43.4 &8.1 \\
&NORA &0.2 &0.0 &0.0 &0.0 &0.0 &0.0 &0.0 &4.6 &0.6 \\
&MV &66.0 &67.0 &\textbf{72.4} &60.2 &54.6 &52.6 &73.8 &72.8 &64.9 \\
&EXP &62.0 &68.2 &\textbf{72.4} &\textbf{63.4} &62.4 &55.6 &73.6 &75.6 &66.6 \\
\rowhighlight
&ICWM (ours)  &\textbf{74.0} &\textbf{70.6} &71.6 &60.4 &\textbf{69.8} &\textbf{62.0} &\textbf{76.6} &\textbf{79.0} &\textbf{70.5} \\
\midrule
\multirow{5}{*}{Long} &$\pi$-FAST &2.6 &0.0 &0.0 &0.0 &0.0 &0.0 &0.0 &2.6 &0.7 \\
&$\pi_{0.5}$ &13.2 &0.0 &0.0 &0.0 &0.0 &0.0 &0.0 &10.0 &2.9 \\
&NORA &1.2 &0.0 &0.0 &0.0 &0.0 &0.0 &0.2 &6.2 &1.0 \\
&MV &36.2 &32.0 &29.6 &24.8 &28.0 &28.2 &33.4 &34.0 &30.8 \\
&EXP &39.8 &32.6 &26.4 &31.0 &30.4 &26.8 &33.2 &39.0 &32.4 \\
\rowhighlight
&ICWM (ours)  &\textbf{45.2} &\textbf{41.2} &\textbf{37.2} &\textbf{39.4} &\textbf{38.4} &\textbf{34.2} &\textbf{39.8} &\textbf{44.6} &\textbf{40.0} \\
\bottomrule
\end{tabular}
\end{table}

\begin{table}[t]\centering
\caption{Success Rates (\%) on Out-of-Domain (Unseen) Viewpoints.}
\label{tab: unseen}
\begin{tabular}{llcccccccc}\toprule
& &$45^\circ$ &$135^\circ$ &$225^\circ$ &$255^\circ$ &$285^\circ$ &$315^\circ$ &Avg \\
\midrule
\multirow{5}{*}{Spatial} &$\pi$-FAST &2.6 &0.0 &0.0 &0.0 &0.4 &3.4 &1.1 \\
&$\pi_{0.5}$ &2.0 &0.0 &0.0 &0.0 &0.0 &8.8 &1.8 \\
&NORA &6.0 &0.0 &0.0 &0.0 &0.0 &3.6 &1.6 \\
&MV &73.2 &\textbf{3.2} &\textbf{18.2} &51.4 &68.2 &\textbf{75.6} &48.3 \\
&EXP &71.2 &1.8 &11.0 &53.6 &65.2 &75.2 &46.3 \\
\rowhighlight
&ICWM (ours) &\textbf{78.2} &2.4 &17.0 &\textbf{55.4} &\textbf{71.2} &75.0 &\textbf{49.9} \\
\midrule
\multirow{5}{*}{Goal} &$\pi$-FAST &0.4 &0.0 &0.0 &0.0 &0.0 &1.0 &0.2 \\
&$\pi_{0.5}$ &29.2 &0.0 &0.0 &0.0 &0.0 &6.2 &5.9 \\
&NORA &0.0 &0.0 &0.2 &0.0 &0.0 &0.0 &0.0 \\
&MV &60.4 &13.2 &13.8 &34.2 &\textbf{53.2} &57.6 &38.7 \\
&EXP &62.6 &\textbf{21.8} &\textbf{18.0} &40.8 &51.2 &54.6 &41.5 \\
\rowhighlight
&ICWM (ours) &\textbf{74.8} &18.6 &\textbf{18.0} &\textbf{41.0} &50.0 &\textbf{62.8} &\textbf{44.2} \\
\midrule
\multirow{5}{*}{Object} &$\pi$-FAST &0.0 &0.0 &0.0 &0.0 &0.0 &4.2 &0.7 \\
&$\pi_{0.5}$ &0.6 &0.0 &0.0 &0.0 &0.0 &6.6 &1.2 \\
&NORA &0.0 &0.0 &0.0 &0.0 &0.0 &0.0 &0.0 \\
&MV &25.4 &\textbf{0.2} &0.4 &11.4 &13.8 &24.8 &12.7 \\
&EXP &23.4 &\textbf{0.2} &0.0 &15.2 &14.8 &\textbf{38.2} &15.3 \\
\rowhighlight
&ICWM (ours) &\textbf{28.6} &0.0 &\textbf{1.0} &\textbf{24.2} &\textbf{20.8} &20.8 &\textbf{15.9} \\
\midrule
\multirow{5}{*}{Long} &$\pi$-FAST &0.0 &0.0 &0.0 &0.0 &0.0 &0.0 &0.0 \\
&$\pi_{0.5}$ &0.4 &0.0 &0.0 &0.0 &0.0 &0.4 &0.1 \\
&NORA &0.0 &0.0 &0.0 &0.0 &0.0 &0.0 &0.0 \\
&MV &30.4 &1.2 &2.8 &24.0 &32.8 &27.6 &19.8 \\
&EXP &29.8 &1.0 &6.4 &24.0 &27.8 &32.2 &20.2 \\
\rowhighlight
&ICWM (ours) &\textbf{36.6} &\textbf{2.2} &\textbf{8.8} &\textbf{28.4} &\textbf{36.6} &\textbf{37.6} &\textbf{25.0} \\
\bottomrule
\end{tabular}
\end{table}

\newpage

\section{Details of Real-world Experiments}
\label{app:real}
\subsection{Experimental Configuration}

\begin{wrapfigure}{r}{0.5\linewidth}
    \vspace{-15pt} 
    \centering
    \includegraphics[width=0.8\linewidth]{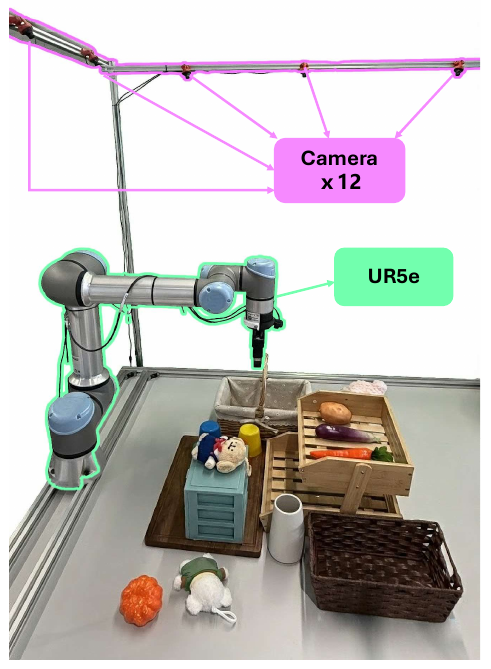}
    \caption{Real-World Experimental Platform. Our physical setup comprises a 6-DoF UR5e manipulator equipped with a parallel gripper and a comprehensive 12-camera perception array distributed at diverse elevations and azimuthal angles.}
    \label{fig:setup}
    \vspace{-20pt} 
\end{wrapfigure}

\paragraph{Physical Platform and Perception.} As illustrated in \autoref{fig:setup}, our real-robot workstation is centered around a 6-DoF UR5e manipulator equipped with a Robotiq 2F-85 parallel gripper, providing a versatile platform for diverse manipulation tasks. To create a challenging multi-view sensing environment, we deploy an array of 12 cameras strategically positioned at varying elevations and azimuthal angles. This setup ensures comprehensive coverage of the workspace while introducing significant perspective-induced spatial distortions.

\paragraph{Generalization Protocol and Data Collection.} To rigorously evaluate zero-shot adaptation, we partition the 12-camera system into two distinct subsets: 6 cameras are designated for training, while the remaining 6 cameras are withheld exclusively for testing. This balanced split ensures that the model is evaluated on a wide range of previously unseen perspectives, requiring success to depend entirely on functional system identification rather than the memorization of specific camera-to-robot geometries. For task-specific knowledge, we collect approximately 100--150 human demonstrations per task via teleoperation, covering four representative manipulation tasks:

1. Spatial Reasoning \& Disambiguation: ``Put the toy on the box into the basket.'' This task requires the agent to understand the vertical spatial relationship between the toy and the box, necessitating precise end-effector positioning to take the toy without disturbing the support surface—a process highly sensitive to viewpoint-induced depth errors

2. Fine-grained Alignment: ``Stack the yellow cup onto the red cup.'' This serves as a benchmark for high-precision motor control, where the agent must align the principal axes of two objects under novel perspective projections.

3. Structural Manipulation: ``Lift the basket.'' This task focuses on handle-centric grasping of large-scale empty containers, testing the model's ability to ground actions on specific structural affordances of an object.

4. Multi-Object Semantic Grounding: ``Pick up the eggplant and place it onto the red plate.'' Conducted in a cluttered scene with multiple distractor objects, this task assesses the model's ability to maintain correct object-instruction alignment when viewed from unfamiliar angles that may cause occlusion or visual overlap.

For task-specific knowledge, we collect approximately 100–150 high-quality human demonstrations per task via teleoperation. These demonstrations capture the necessary precision and coordination for successful execution across diverse initial object layouts.

\subsection{Details of Generalization Experiments}
\label{app:gen}
\textbf{Semantic Perturbations.} 
Semantic perturbation experiments are evaluated across 4 tasks × 4 in-domain viewpoints × 10 trials per condition (160 trials per bar). We evaluate two types of scene-level variation: (1) Distractor objects: 10 task-irrelevant objects are placed within the workspace; (2) Novel table textures: the table surface is replaced with 4 unseen textures absent during training.

\textbf{Morphological Generalization-UR5E.} 
Morphological generalization experiments are evaluated across 4 tasks × 4 in-domain viewpoints × 10 trials x 3 morphological settings, with rigid spacers of $\Delta L \in \{20, 40, 80\}$ mm attached to the gripper flange to systematically alter forward kinematics at test time.

\begin{figure}[t]
    \centering
    \includegraphics[width=\linewidth]{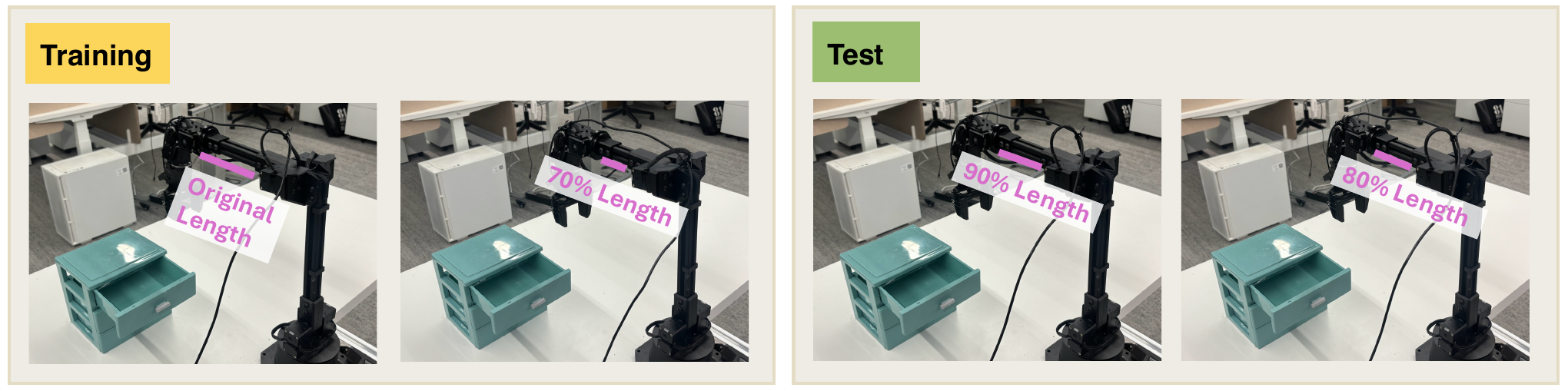}
    \caption{Additional morphological generalization evaluation on the WindowX platform. We shorten the robot's link lengths to \{100\%, 90\%, 80\%, 70\%\} of the original, yielding four distinct morphological configurations.}
    \label{fig:window}
\end{figure}

\textbf{Morphological Generalization-WindowX.} 
\label{app:morph_windowx}
To further validate ICWM's generalization capability beyond rigid spacer attachments, we conduct an additional experiment on a WindowX platform with systematically varied link lengths. Specifically, we scale the robot's link lengths to \{100\%, 90\%, 80\%, 70\%\} of the original, yielding four distinct morphological configurations (\autoref{fig:window}). 
The model is trained on the two boundary configurations (100\% and 70\% link length) and evaluated on the two interpolated out-of-distribution configurations (90\% and 80\%), assessing generalization to unseen kinematic structures without any parameter updates. Experiments are conducted across 3 tasks $\times$ 20 trials $\times$ 2 morphological settings (120 trials total).

\section{Details of Self-Exploration Probing}
\label{app:random}
\subsection{Implementation Details}

To enable test-time adaptation, we implement a stochastic self-exploration phase that allows the agent to synchronize its internal coordinate system with the current camera viewpoint. This phase is designed to be task-agnostic and prioritizes safety during real-robot deployment.

The exploration process involves moving the robot's end-effector toward a randomly sampled target coordinate $g_{probe} \in \mathbb{R}^3$. The control is executed via a fixed-step movement policy rather than a goal-reaching controller. It is important to note that the robot is not required to reach the target point; the primary objective is to generate diverse visual-motor transitions $(o_t, a_t, o_{t+1})$ that expose the underlying action-observation mapping under the current configuration $\psi$. These interaction fragments are then collected and organized as the in-context prefix $\mathcal{T}$.

\begin{figure}[!t]
    \centering
    \includegraphics[width=0.9\linewidth]{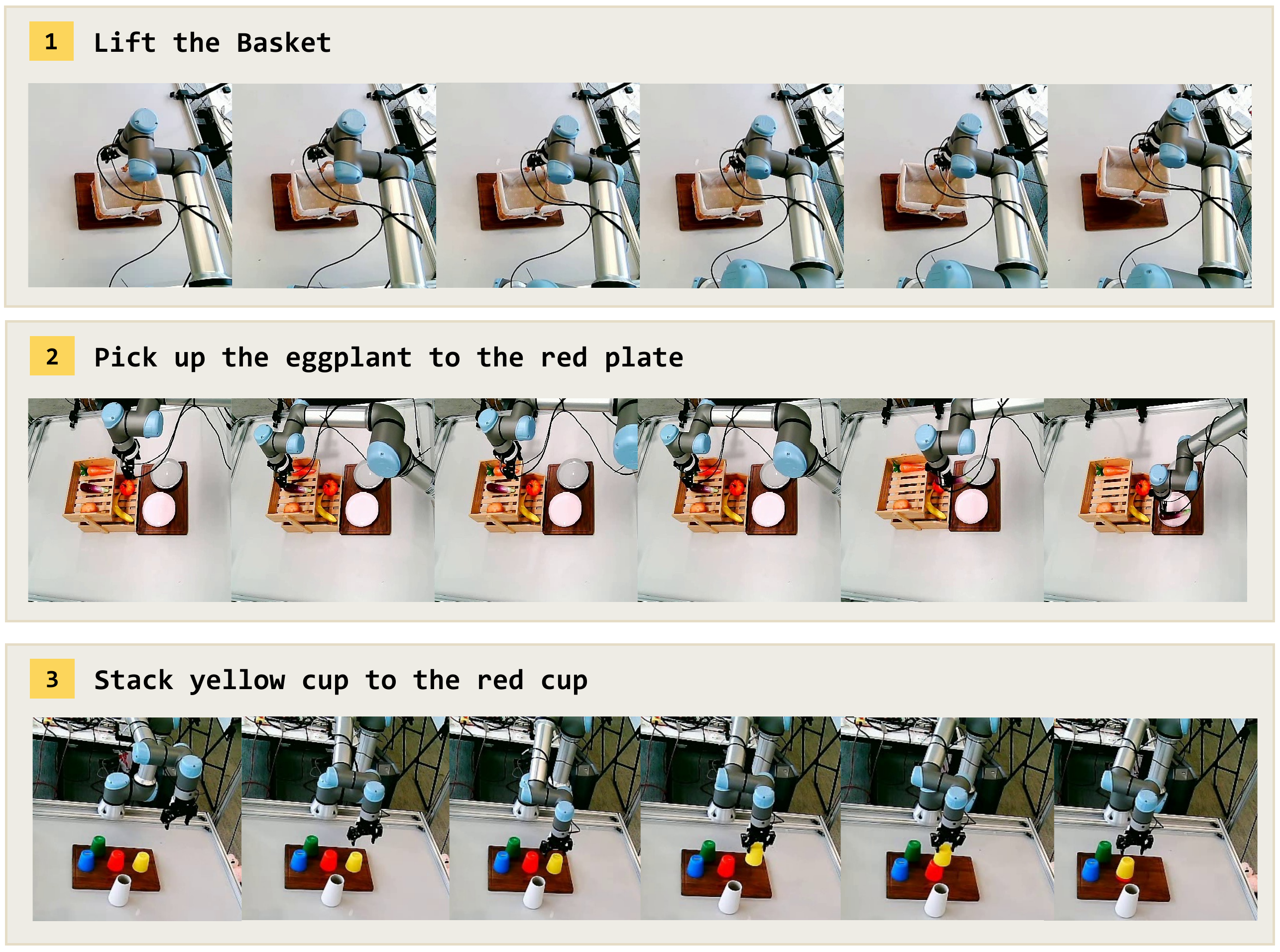}
    \caption{The supplementary successful trajectories from real-robot experiments. These trajectories illustrate the ICWM model's capability in completing lifting and pick-and-place maneuvers, as well as its robustness against object disturbances.}
    \label{fig:realcase2}
\end{figure}

\textbf{Workspace-Constrained Random Exploration (Real-Robot).} For real-world experiments, probing targets $g_{probe}$ are sampled uniformly at random within the robot's reachable workspace.
Rather than requiring manual specification, this workspace bounding box is derived automatically from the robot's forward kinematics model—given the joint limits of the UR5e, the reachable Cartesian space in the robot base frame can be computed analytically, requiring no human annotation or scene-specific calibration.
The robot moves toward each sampled target via fixed-step increments, ensuring all motions remain within the operable joint limits. The resulting transitions cover diverse spatial regions of the workspace, forming the in-context prefix $\mathcal{T}$.

Crucially, since the bounding box is defined entirely in the robot's base frame, it remains invariant across arbitrary camera viewpoint shifts or semantic scene changes, requiring only a one-time specification for a fixed physical workstation setup. The random movements are task-agnostic and configuration-agnostic, carrying no task-relevant information. Our zero-shot claim refers specifically to the absence of task-specific demonstrations or parameter updates for any novel configuration; in simulation, this step is unnecessary entirely.

\textbf{Timing Analysis.}
The probing phase introduces minimal overhead. Before task execution, the robot performs 20 probing actions spanning the workspace in approximately 5--6 seconds total. During task execution, $N{=}5$ triplets are randomly sampled from these 20 recorded transitions as the in-context prefix at each inference step. Crucially, this 20-step probing is performed \emph{once per deployment}: the same context pool is reused throughout the entire task execution with no additional probing required. Given the multi-step nature of VLA manipulation tasks, this one-time overhead constitutes a small fraction of total deployment time, and is further amortized to negligible levels when the same configuration is reused across tasks.

\subsection{Case Visualization}

\begin{figure*}
    \centering
    \includegraphics[width=1\linewidth]{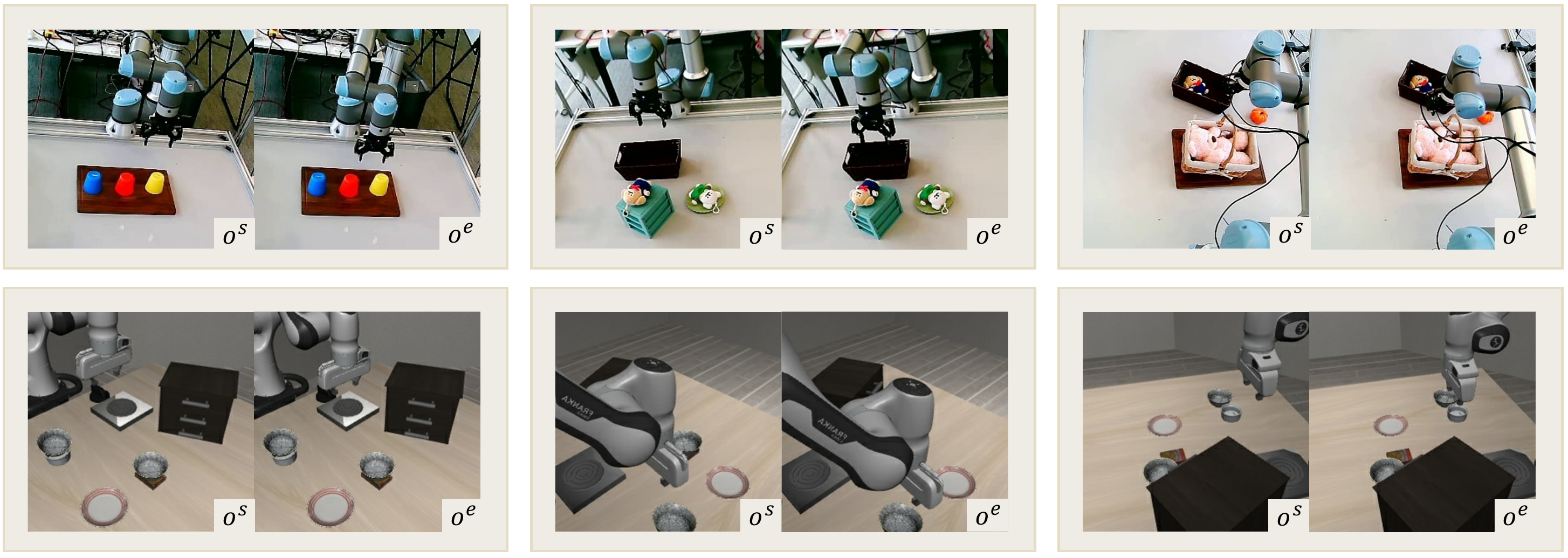}
    \caption{Visualization of start ($o^s$) and end frames ($o^e$) for random interaction clips in the real world and simulation.}
    \label{fig:random}
\end{figure*}

We present the start ($o^s$) and end frames ($o^e$) of random action interaction clips from both the real world and simulation in \autoref{fig:random}.

\end{document}